\def\year{2022}\relax
\documentclass[letterpaper]{article} 
\usepackage{lipsum}
\usepackage{amsmath}
\usepackage{amsthm}
\usepackage{amssymb}
\usepackage{amsfonts}
\theoremstyle{definition}
\newtheorem{definition}{Definition}
\usepackage{mathrsfs}
\usepackage[english]{babel}
\usepackage[autostyle]{csquotes}

\usepackage{aaai22}  
\usepackage{times}  
\usepackage{helvet}  
\usepackage{courier}  
\usepackage[hyphens]{url}  
\usepackage{graphicx} 
\usepackage{natbib}  
\usepackage{caption} 
\DeclareCaptionStyle{ruled}{labelfont=normalfont,labelsep=colon,strut=off} 
\usepackage{algorithm}
\usepackage{algorithmic}

\usepackage{newfloat}
\usepackage{listings}
\floatstyle{ruled}
\newfloat{listing}{tb}{lst}{}
\floatname{listing}{Listing}
\title {Modelling Control Arguments via Cooperation Logic in Unforeseen Scenarios}
\author{
    Minal Suresh Patil
}

\affiliations{
    Umeå universitet\\

    UNIVERSITETSTORGET 4\\
    907 36 Umeå, Sweden\\
    minal.patil@umu.se
}

\usepackage{bibentry}

\begin{document}

\maketitle

\begin{abstract}
The intent of control argumentation frameworks is to specifically model strategic scenarios from the perspective of an agent by extending the standard model of argumentation framework in a way that takes unquantified uncertainty regarding arguments and attacks into account. They do not, however, adequately account for coalition formation and interactions among a set of agents in an uncertain environment. To address this challenge, we propose a formalism of a multi-agent scenario via cooperation logic and investigate agents' strategies and actions in a dynamic environment.
\end{abstract}

\section{Introduction}

Humans must possess beliefs in order to engage with their surrounding environments successfully, coordinate their activities, and be capable of communicating. Humans sometimes use arguments to influence others to act or realise a particular approach, to reach a reasonable agreement, and to collaborate together to seek the optimal possible solution to a particular problem. In light of this, it is not unexpected that many recent efforts to represent artificially intelligent agents have incorporated arguments and beliefs of their environment. Argumentation-based decision-making approaches are anticipated to be more in line with how people reason, consider possibilities and achieve objectives. This confers particular advantages on argumentation-based techniques, including transparent decision-making and the capability to provide a defensible rationale for outcomes. In our recent work, we propose the use of explanations in autonomous pedagogical scenarios~\cite{patil2022explainability}. That is, how explanations should be tailored in multi-agent systems(MAS) (teacher-learner interaction).
In many domains, including legal reasoning and consensus-building, formal argumentation has been utilised to provide a natural and simple form of non-monotonic reasoning~\cite{rahwan2009argumentation, bistarelli2013coalitions, bench2007argumentation, kakas2014argumentation}. Abstract models of argumentation play an important part in studies aimed at developing a formal model for argument-based inference because they offer an answer to a core issue: how should a rational agent choose among a conflicting set of more justified arguments? The term \enquote{abstract} emphasises how well these models overlook the nature and structure of arguments in support of emphasising on the possible semantics that could be utilised to provide a coherent response to the initial question. The most typical abstract formulation of argumentation is expressed by directed graphs, which Dung originally developed in ~\cite{dung1995acceptability} as argumentation frameworks (AFs).

\noindent Control argumentation frameworks (CAFs)~\cite{dimopoulos2018control} are a new class of formal argumentation that aims to achieve a certain state where acceptance or rejection of specific arguments is guaranteed, notwithstanding any unpredictable challenges they may confront in a dynamic environment expressed by a set of uncertain arguments. Control argumentation frameworks (CAFs) are a new class of formal argumentation that aims to achieve a certain state where acceptance or rejection of specific arguments is guaranteed, notwithstanding any unpredictable challenges they may confront in a dynamic environment expressed by a set of uncertain arguments. By allowing for the specification of control (C), uncertain (U), and fixed arguments (F) that result in uncertainty regarding the attack relation, CAFs generalise AFs from a syntactic standpoint. CAFs have already been demonstrated to be a natural way to approach argumentation-based negotiation in scenarios where knowledge about the opponent's profile is insufficient.~\cite{dimopoulos2019argumentation} The primary objective of CAFs is to investigate how to uphold the desired status of specific arguments that can withstand unforeseen circumstances of the uncertain arguments. 

\noindent Present studies of CAFs, however, may not be immediately applicable to multi-agent scenarios since they merely capture the statues of few arguments as opposed to how agents interact with one another within an uncertain environment. In a CAF, it is presumed all agents are aware of the entire framework, whilst in practice, agents could come across certain unexpected circumstances in their environment. 

To give a practical example, autonomous Intent-Based Networking (IBN)~\cite{campanella2019intent} captures and translates business intent into network policies that can be automated and applied consistently across the network. The goal is for the network to continuously monitor and adjust its performance to assure the desired business outcome. Intent allows the agent to understand the global utility and the value of its actions. Consequently, the autonomous agents can evaluate situations and potential action strategies rather than being limited to following instructions that human developers have specified in policies. In these cases, agents may adjust their model of the environment as well as their strategy according to information provided by the environment. There are several other circumstances in which the agent may not be able to guarantee a specific status of specific arguments and would necessitate assistance from other agents. Agents may not always know the optimal strategy until they form a coalition. The following area of investigation naturally arises because CAFs fall short of encapsulating these subtleties:

\textit{How, in the context of CAFs, can agents update their beliefs, strategies, and coalitions in response to unintended shifts in the environment?}

\noindent We provide an approach based on cooperation logic and transition systems to investigate this issue, and we utilise a formal language and semantics based on the prior theory of alternating-time temporal logic (ATL) to express coalition propositions and formations in the context of CAFs. The following outline provides an overview of the layout of this research. In the next part, we offer a concise introduction to some fundamental concepts about argumentation frameworks, specifically the control argumentation framework and cooperation logics.




\section{Background}
\subsection{Argumentation Framework}

\noindent We provide a brief formal overview of Dung's framework because it is one of the most well-known settings for abstract argumentation.



\noindent An Abstract Argumentation Framework $(A A F)$ $F$ is a pair $\langle A, D\rangle$, where $A$ is a finite and non-empty set, whose elements are called \textit{arguments}, and $D \subseteq A \times A$ is a binary relation over $A$, whose elements are called \textit{attacks}. The graph having $A$ and $D$ as set of nodes and edges, respectively, is called \textit{argumentation graph} of $F$. Given $a, b \in A$, we say that $a$ \textit{attacks} $b$ iff $(a, b) \in D$. A set $S \subseteq A$ \textit{attacks} an argument $b \in A$ iff there is $a \in S$ that \textit{attacks} $b$. An argument $a$ attacks $S$ iff $\exists b \in S$ attacked by $a$.

A set $S \subseteq A$ of arguments is said to be \textit{conflict-free} if there are no $a, b \in S$ such that $a$ attacks $b$. An argument $a$ is said to be \textit{acceptable} w.r.t. $S \subseteq A$ iff $\forall b \in A$ such that $b$ \textit{attacks} $a$, there is $c \in S$ such that $c$ \textit{attacks} $b$. 

An \textit{extension} is a set of arguments that is considered "reasonable" according to some semantics. In particular, we consider the following semantics from the literature:

\begin{itemize}
    \item admissible $(a d): S$ is an admissible extension iff $S$ is conflict-free and its arguments are acceptable w.r.t. $S$;
    \item stable (st): $S$ is a stable extension iff $S$ is conflict-free and $S$ attacks each argument in $A \backslash S$;
    \item complete $(\mathrm{co}): S$ is a complete extension iff $S$ is admissible and every argument acceptable w.r.t. $S$ is in $S$;
    \item grounded (gr): $S$ is a grounded extension iff $S$ is a minimal (w.r.t. $\subseteq)$ complete set of arguments;
    \item preferred (pr): $S$ is a preferred extension iff $S$ is a maximal (w.r.t. $\subseteq)$ complete set of arguments;
\end{itemize}

\textit{Accepted arguments.} An argument $a$ is (credulously) accepted under a semantics $\sigma$ iff $a$ belongs to some $\sigma$ extension of $F$. In some sense, checking the acceptability of an argument is a way of deciding whether $a$ represents a robust point of view in the discussion modeled by $F$.

\subsection{Control Argumentation Framework}
Now, we introduce control argumentation frameworks~\cite{dimopoulos2018control}. Besides the classical arguments and attacks, a $CAF$ is made of 1) uncertain information (that can represent the information that an agent has about the environment or about the other agents) and 2) control arguments (that represent the possible ways for the agent to reason with the environment or the other agents). Formally speaking, $CAF=\langle F, C, U\rangle$ describes a model of the world or environment that comprises arguments that are managed by the agent $(C)$ i.e. the agent can utilise to counteract attacks from the uncertain component's arguments in the fixed part $F$, arguments that are guaranteed but not controlled $(F)$, and uncertain arguments that may appear in the system, but could be absent $(U).$ The uncertain component pertains to elements that, depending on how a dynamic environment evolves or how other agents interact, if it is a part of the system (or not). 

A \textit{CAF} is a triple $CAF=\langle F, C, U\rangle$ where:

\begin{itemize}

\item $F=\left(A^{F}, R^{F}\right)$ is the \textit{fixed part}, with $R^{F} \subseteq\left(A^{F} \cup \Bar{A}\right) \times\left(A^{F} \cup \Bar{A}\right)$, and both $A^{F}$ and $\Bar{A}$ are two finite sets of arguments;

\item $C=\left(A^{C}, R^{C}\right)$ is the \textit{control part} where $A^{C}$ is a finite set of arguments and $R^{C}\subseteq\left(A^{C} \times\left(A^{F} \cup \Bar{A} \cup A^{C}\right)\right) \cup \left(\left(A^{F} \cup \Bar{A} \cup A^{C}\right) \times A^{C}\right);$

\item $U=\left(\Bar{A},\left(\Bar{R} \cup R^{\leftrightarrow}\right)\right)$ is the \textit{uncertain part}, where $$\Bar{R}, R^{\leftrightarrow} \subseteq\left(A^{F} \cup \Bar{A}\right) \times\left(A^{F} \cup \Bar{A}\right)$$

$A^{F}, \Bar{A}$, $A^{C}$; $R^{F}, \Bar{R}, R^{\leftrightarrow}$, and $R^{C}$ are disjoint subsets and $R^{\leftrightarrow}$ is symmetric and irreflexive.\footnote{with no loss of generality both the assumptions are valid\cite{niskanen2021controllability}}
\end{itemize}

The objective is to identify an action, specifically a portion of the arguments within the agents' control, which guarantees a specified status for given arguments in all scenarios and we can. If this is guaranteed, we say that the status is controlled by the strategy or action. In a MAS setting introduced in the next section, we use a \textit{coalition proposition} to represent the controlled status of a certain argument with respect to an agents' state, which can be denoted as $\zeta(a)$ as accepting an argument $a$.

\subsection{Cooperation logics and Transition Systems}
Cooperation logics such as Alternating-time Temporal Logic (ATL)~\cite{alur2002alternating} and Coalition Logic (CL)~\cite{pauly2001logic} have been shown to be powerful, effective and intuitive knowledge representation formalisms for such games, and game theoretic models of cooperation have proven to be a useful source of methods and insights for the field of MAS. In this work, we mainly aim to focus on ATL, and although it shares some similarities with Coalition Logic, ATL was developed with entirely separate ambitions and emerged from a very different research community. ATL has been used as a formal system for the specification and verification of MAS. Examples of such work include formalising the notion of role using ATL~\cite{ryan2001agents}, the development of epistemic extensions to ATL~\cite{pauly2003logic}, and the use of ATL for specifying and verifying cooperative multi-agent behaviour~\cite{van2005logic}.


\section{Our Approach}
In this section, we introduce the notion of a multi-agent scenario set in the context of a control argumentation framework where agents can form coalitions and update their actions/strategies and the beliefs of their environment.  We extend the existing ATL framework, which will incorporate CAF and ATL (CAFATL). First, we present the original semantic structures over which the ATL formulae are interpreted in order to provide a precise description of CAFATL. 

\begin{definition}[ATS]

\noindent An \textit{alternating transition system (ATS)} is a 5-tuple $S=\langle\Pi, \Sigma, Q, \pi, \delta\rangle$, where:

\begin{itemize}
    \item $\Pi:$ represent a finite, non-empty set of Boolean variables. 
    \item $\Sigma=\left\{a_{1}, \ldots, a_{n}\right\}$ represent a finite, non-empty set of agents. 
    \item $Q:$ represent a finite, non-empty set of states.
    \item $\pi: Q \rightarrow 2^{\Pi}$ represent the set of Boolean variables satisfied in each state.
    \item $\delta: Q \times \Sigma \rightarrow 2^{2^{Q}}$ is the system transition function, which maps a state and an agent to a non-empty set of choices, where each choice is a set of possible next state.
\end{itemize}

\end{definition}

Here, we formalise the semantic structure CAFATL, by modifying the original ATS as follows:

\begin{definition}[$CAFATL$]

\noindent A $CAF$ based Multi-Agent Coalition Transition System is a 8-tuple $S=(\Pi, \Sigma, Q, \pi, \delta, \Theta, K, \Upsilon)$, where:

\begin{itemize}
    \item $\Pi:$ represent a finite, non-empty set of propositions.
    \item $\Sigma=\left\{a_{1}, \ldots, a_{n}\right\}$ represent a finite, non-empty set of agents. 
    \item $Q:$ represent a finite, non-empty set of states.
    \item $\pi: Q \rightarrow 2^{\Pi}$ represent the set of propositions that are satisfied/true in each state.
    \item $\Theta: Q \times \Sigma \rightarrow \mathbb{N}$ is a system function which assigns a numerical value to each available action or strategy for a particular agent $i \in \Sigma$ and for a state $q \in Q$ where $\mathbb{N}$ is a set of natural numbers. For every state $q$, we identify the set of coalition strategies/actions for every agent in $\Sigma$ by $\Sigma(q).$
    \item $\delta: Q \times \Sigma(q) \rightarrow 2^{2^{Q}}$ is the system transition function, in which $\delta(q, \tau)$ represents the next state from $q$ if the agent executes the strategy $\tau \in \Sigma(q)$. 
    \item $K=\left\{\kappa_{0}, \ldots, \kappa_{n}\right\}$ represent the set of all possible control argumentation frameworks the agent encounters and $\kappa_{0}$ represent the control argumentation framework for the main dynamics of the environment in which the coalition proposition is determined. 
    \item $\Upsilon: K \times \Sigma(q) \rightarrow \Sigma $   is a system function which updates the agents’ models of the environment where $\Upsilon(\kappa_i, \tau)$ represents the new updated agents' model from the former model $\kappa_i$ when a coalition strategy $\tau \in \Sigma(q)$  is carried out. 
\end{itemize}
\end{definition}

Further, we formalise a language representing the coalition propositions for the above multi-agent scenario and are mainly concerned with the following characteristics:

\begin{itemize}
    \item if the status of a certain argument is controlled at a specific agent state.

    \item if a coalition proposition can be semantically entailed with some other coalition proposition.

    \item if the status of a particular argument can be governed by a any coalition of the agents.
\end{itemize}

\begin{definition}[Formal Language $\mathscr{L}$]

\begin{equation}
    p \in \Pi|\neg \phi| \phi \wedge \psi|\phi \rightarrow \psi|\langle\langle \mathfrak{C}\rangle\rangle \phi
\end{equation}

\begin{itemize}
    \item $q \vDash p$ iff $p \in \pi(q)$
    \item $q \vDash \neg \phi$ iff $q \not \models \phi$
    \item $q \vDash \phi \wedge \psi$ iff $q \vDash \phi$ and $q \vDash \psi$
    \item $q \vDash \phi \rightarrow \psi$ iff for all states $q^{\prime} \in Q$, if $q^{\prime} \vDash \phi$ then $q^{\prime} \vDash \psi$
    \item $q \langle\langle \mathfrak{C}\rangle\rangle \phi$ iff there exists a set of agents $\mathfrak{C} \subseteq \Sigma$ and a joint strategy $a_\mathfrak{C}$ of $\mathfrak{C}$ such that for any $\tau \in \Sigma(q)$, if $a_\mathfrak{C} \subseteq \tau$ then $\delta(q, \tau) \vDash \phi$
\end{itemize}
For any $\mathcal{P} \subseteq \Pi$, we write $q \vDash \mathcal{P}$ for $q \vDash \bigwedge_{a \in \mathcal{P}} a$

\end{definition}
By formalising the multi-agent scenario and its corresponding formal language and semantics, we are capable of representing coalition propositions of  multi-agents.



\section{Conclusion}
In this work, we present a multi-agent scenario in the context of CAFs where agents’ interact with one another and the environment and simultaneously update their beliefs and actions. Ongoing work is focused on characterising a preference relation in which agents can assign specific values (e.g.societal values) to each coalition. It is a reasonable way to describe and reason about the relative strength of arguments; hence, we consider the notion of preference essential. For instance, instead of merely adding new arguments to agents’ models, we may identify them as fixed arguments or arguments that a specific agent controls. The
fixed arguments and those influenced by other agents or a group of agents must be recognised by an agent in the context of control argumentation. Therefore, modelling preferences for arguments that other agents influence can facilitate, for example, deception detection/(mis)trust building between agents.


\section{Acknowledgments}
This work was partially supported by the Wallenberg AI, Autonomous
Systems and Software Program (WASP) funded by the Knut and
Alice Wallenberg Foundation. 


\end{document}